\documentclass{ifacconf}

\usepackage{color}
\usepackage{natbib}            
\usepackage{graphicx}          
\usepackage{multirow}				
\usepackage{amsmath}

\usepackage[breakable, theorems, skins]{tcolorbox}
\tcbset{enhanced}

\usepackage{tikz}
\usetikzlibrary{arrows, positioning, trees,matrix, calc}

\DeclareRobustCommand{\mybox}[2][gray!20]{%
\begin{tcolorbox}[   
        breakable,
        left=0pt,
        right=0pt,
        top=0pt,
        bottom=0pt,
        colback=#1,
        colframe=#1,
        width=8.8cm, 
        enlarge left by=0mm,
        boxsep=5pt,
        ]
        #2
\end{tcolorbox}
}

\begin{document}

\begin{frontmatter}

\title{Online Vision- and Action-Based Object Classification Using Both Symbolic
and Subsymbolic Knowledge Representations}
\author[First]{Laura Steinert}
\author[First]{Jens Hoefinghoff}
\author[First]{Josef Pauli}

\address[First]{Universit\"at Duisburg-Essen, Fakult\"at f\"ur
Ingenieurwissenschaften, Bismarckstrasse 90, 47057 Duisburg (e-mail:
laura.steinert@stud.uni-due.de; jens.hoefinghoff@uni-due.de;
josef.pauli@uni-due.de)}


\begin{abstract}                          
If a robot is supposed to roam an environment and interact with objects, it is often
necessary to know all possible objects in advance, so that a database with models of all objects can be generated for visual identification. However, this constraint cannot always be fulfilled. Due to that reason, a model based object recognition cannot be used to guide the robot’s interactions.
Therefore, this paper proposes a system that analyzes features of encountered objects and then uses these features
to compare unknown objects to already known ones. From the resulting similarity appropriate actions can be derived. 
Moreover, the system enables the robot to learn object categories by grouping similar objects or by splitting existing categories. To represent the knowledge a hybrid form is used, consisting of both symbolic and subsymbolic representations.
\end{abstract}

\end{frontmatter}

\section{Introduction}

At the beginning of the 21st century robots start to be part of the society and act in e.g. private households. However, performing tasks in a strictly controlled laboratory and performing the same tasks in the \emph{real} world are two completely different things. Most of all, the environment is much more unpredictable, dynamic and complex. Yet when robots are supposed to help humans physically in their everyday life, this problem needs to be approached.

In order to interact with their environment robots often need a database for visual identification that contains models of all objects they can encounter. However, for such unspecified environments as human households a database containing models of all possible objects is not feasible. However, if a robot is supposed to learn patterns of interactions with its environment, object identification is essential. 
To overcome this problem, a system is proposed that is able to learn objects online based on abstract features. Additionally, it is able to group these objects into categories to predict suitable interaction patterns for new objects by comparing them to its learned knowledge base.

The problem of knowledge generalization as well as analogy finding is hardly new and has been studied before. An example system for analogy making is \emph{COPYCAT} by \cite{Mitchell}. \emph{COBWEB} by \cite{Fisher} on the other hand is an example for knowledge generalization.

Yet it is the ability to generalize and to make analogies that seem to be at the core of human intelligence. As such it is a subject that greatly concerns artificial cognitive systems. The presented approach consists of a hybrid knowledge representation using both symbolic and subsymbolic representations. Additionally, it takes the so called \emph{symbol grounding problem} by \cite{Harnad:90} into account. The presented system has been evaluated using a NAO robot of the Aldebaran company.

At the beginning of this paper the terms symbolic, subsymbolic and hybrid knowledge representation as well as the symbol grounding problem are defined. Afterwards, an example scenario is presented that is used in the remaining paper. This is followed by a section that describes the actual framework, first the subsymbolic, then the symbolic parts. At the end of the paper, the evaluation of the framework and the results are discussed before a conclusion is given.

\section{Knowledge Representation and Symbol Grounding}

The decision on a specific knowledge representation is always influenced by the task to be solved. When using a \emph{symbolic knowledge representation}, the knowledge is encoded explicitly. Examples for such representations are graphs. Symbolic representations are well suited to represent relations between objects (cf. \cite{French}), for instance to model natural language (cf. \cite{Harnad:90}). 

\emph{Subsymbolic representations} on the other hand encode knowledge implicitly, as it is done in neural networks (cf. \cite{Luger}). Such a representation can be good for learning motoric tasks or for analyzing sensory data (cf. \cite{Harnad:90}). 

Both symbolic and subsymbolic representations seem to be used by the human brain. Therefore, artificial intelligence should incorporate both representations and combine them into a \emph{hybrid representation} (cf. \cite{Kokinov:94}). This view has started to be incorporated into newer artificial intelligence systems (cf. \cite{Harnad:90}, \cite{Mitchell}). However, since knowledge represented symbolically can be very abstract, the risk is high that it is no longer connected to the underlying sensory data and is therefore ungrounded. This leads to the symbol grounding problem.

Stevan \cite{Harnad:90} coined the term \emph{symbol grounding problem} by questioning whether a robot really knows what it seems to know. This directly refers to the problem of symbolically represented knowledge being isolated from sensory data. If a robot knew that \emph{dogs can be petted}, it were of hardly any practical use if the robot were not able to recognize a dog standing in front of it. 

The presented approach tries to overcome this problem by embedding the sensory information directly into the knowledge base.

\section{Example scenario}
\label{sec:example}

Before the actual framework is presented, an example scenario is given that is used throughout the paper.
The robot is presented apples of either green, red or brown color sequentially, as well as wooden rectangular toy blocks that are either green, red or yellow. It is then asked to verbally sort the objects into a toy box, a fruit basket and a rubbish bin. After every action the robot is either positively or negatively rewarded. For every encounter with an object only one action can be chosen. If the robot sorts green or red apples into the fruit basket, it receives a positive reward. The same holds for sorting wooden toy blocks into the toy box and for putting brown apples into the rubbish bin. For all other actions the robot receives a negative reward. At the beginning of the scenario the robot has no knowledge regarding any of the objects.

The expectation is that the robot will create three object categories -- one for both green and red apples, one for brown apples and one for all wooden toy blocks.

This scenario seems to be predestined to be solved by a neural network. But while an artificial neural network often needs to be trained anew whenever the scenario changes slightly, the presented system can adapt to such changes by itself. An example of such a change is the introduction of a new object category.

\section{Framework}

In this section the framework is presented. First, a brief overview of the system is given and the basic underlying algorithm is explained. After that, another subsection describes the feature extraction, which is one of two major components of the system. A subsection regarding the knowledge management unit, the second major component of the system, follows.

\subsection{Overview}

\begin{figure}
\begin{center}
\includegraphics[angle=0,width=8.4cm]{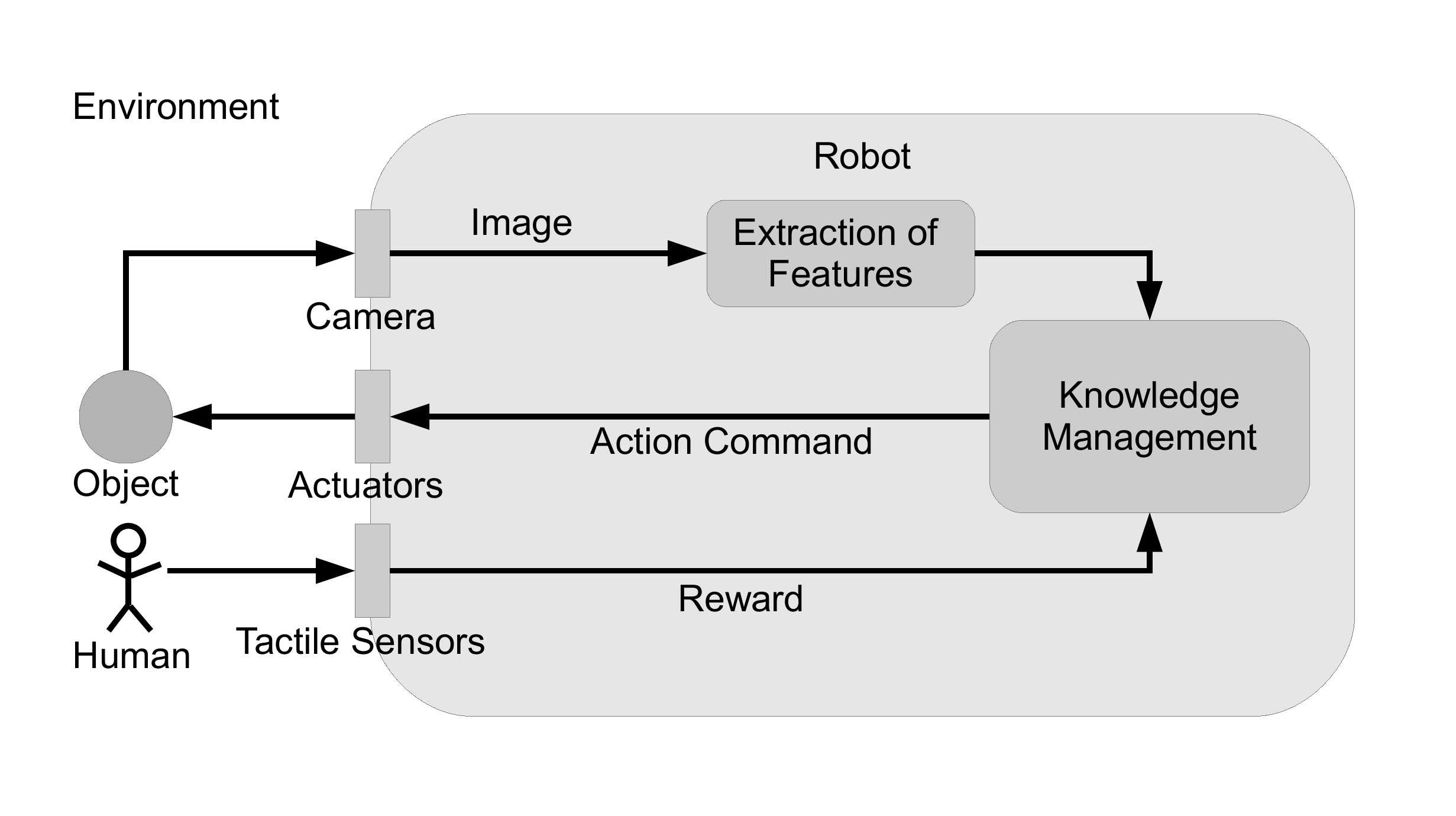}    
\caption{Overview of the system’s parts and their interactions with each other and the environment}  
\label{fig:wholeSystem}                                 
\end{center}                                 
\end{figure}

The system presented uses a hybrid knowledge representation as explained above. As such it can be divided into two sections: one that uses a symbolic and one that uses a subsymbolic representation of knowledge. Figure \ref{fig:wholeSystem} shows an overview of the system. 

Whenever an object is presented to the robot, an image of the object is taken. Based on this image, features are extracted -- in the case of the example scenario color and basic shape (rectangular or circular). The feature extraction components -- one for each feature -- can be multithreaded and independent of each other. In these components a subsymbolic representation is used. Furthermore, the usage of other sensors to extract the feature set is also possible. 

\begin{figure}
\begin{center}
\includegraphics[angle=0,width=6.0cm]{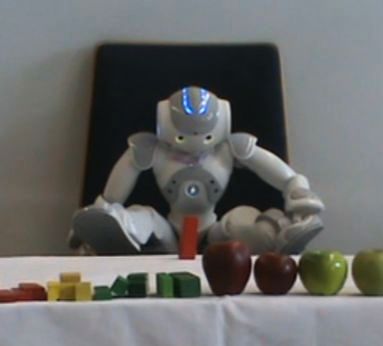}    
\caption{A red wooden toy block is shown to the robot}  
\label{fig:nao}                                 
\end{center}                                 
\end{figure}

Figure \ref{fig:nao} shows the robot while performing the example scenario. At the displayed time it is shown a red wooden toy block. The foreground of the image shows other objects used in the scenario.

After the feature set is computed, it is relayed to the knowledge management, which uses a symbolic representation for its knowledge base. The robot first checks whether an object that fits the currently perceived feature description has been encountered before. If this is the case, an action based on the previous experience with this object is chosen. If this object has not been seen before, it is compared to all known object categories. Depending on the similarity to other objects, an appropriate action is chosen. In both cases the chosen action is performed. Afterwards the robot is given a reward for the chosen action by a human supervisor. In the implemented version of the system this was accomplished by using the tactile sensors of the robot. The reward is then used to update the knowledge base, e.g. by combining or dividing object categories.

The two following subsections describe in greater detail the symbolic and the subsymbolic parts of the system.

\subsection{Feature extraction}

The feature extraction components use a subsymbolic knowledge representation. In the case of the example scenario the features \emph{color} and \emph{basic shape} have been extracted from images using neural networks. This feature set alone is of course not sufficient for a robot to interact in a human household. Yet it is sufficient to evaluate the system.

For each of the two features to be extracted, a separate multilayer perceptron with one hidden layer is used. Since the example scenario is executed under natural lighting conditions, color variations stemming from changing light conditions are to be expected. In order to minimize this effect, the L*a*b color space is used and the L-channel -- which contains the luminacity data -- is omitted. 

\begin{figure}
\begin{center}
\includegraphics[angle=0,width=8.4cm]{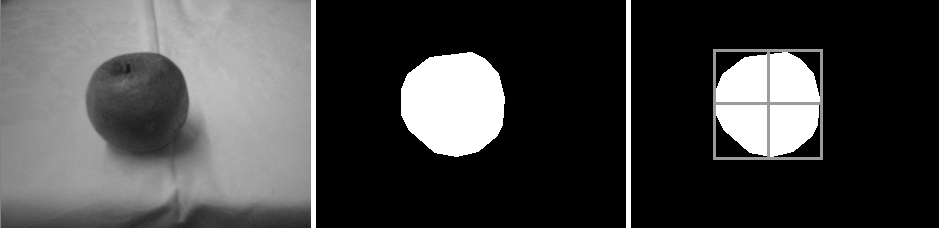} 
\caption{From left to right: The original image (converted to black and white), the perceived object silhouette and the quartered bounding box}
\label{fig:colourextraction}
\end{center}
\end{figure}

Yet before the color is analyzed, a preprocess identifies the contour of the object in the image. Afterwards, a bounding box is formed around the object. Figure \ref{fig:colourextraction} shows a binary image of an object, the silhouette of the detected object and a quartered bounding box superimposed on the silhouette.
For every quarter of the bounding box, the average color values of the pixels inside the object contour in the original colored image are then computed. 

\mybox{As an example, consider the following matrix containing the object's color values of the b-channel of the upper right quarter of the bounding box. The background of the image, i.e. everything of the image that does not depict the object, was replaced by zeros. Therefore, all non-zero values encode pixels of the object.

\[\begin{matrix}
	0 & 0 & 0 & 0 & 0\\
	130 & 0 & 0 & 0 & 0\\
	134 & 137 & 0 & 0 & 0\\
	138 & 135 & 139 & 0 & 0\\
	140 & 138 & 140 & 0 & 0\\
\end{matrix}\]

For each of such quarters and for both the a- and b-channel values the average of all pixels with values greater than zero is computed. In the example case this is approximately 137.}

The resulting eight values are used as an input for the multilayer perceptron, classifying the color to be red, green, yellow or brown. The classification values are then converted to percentages.

The used multilayer perceptron has one hidden layer consisting of ten nodes and was trained with the \emph{RPROP} (resilient backpropagation) algorithm.  

In order to analyze the form of an object, the object contour is extracted first. Afterwards, another bounding box with varying orientation as well as the minimal enclosing circle are computed. Then the area of the bounding box, the circle and the object itself are determined. The relation of the area of the object to that of the bounding box as well as the relation of the area of the object to that of the circle are then used as an input for the multilayer perceptron analyzing the shape. 
This artificial neural network has one hidden layer with two nodes and was trained with the \emph{RPROP} algorithm, too.

\begin{figure}
\begin{center}
\includegraphics[angle=0,width=6.0cm]{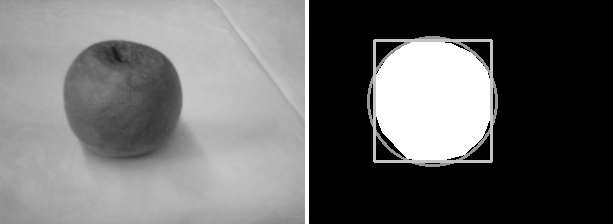} 
\caption{The original image (left) and the perceived object silhouette together with the bounding box and minimal enclosing circle}
\label{fig:formextraction}
\end{center}
\end{figure}

Figure \ref{fig:formextraction} displays the necessary steps before the actual form extraction takes place. The left-hand side of the figure shows the original image. On the right, the object silhouette as well as the bounding box and minimal enclosing circle are shown.

The resultant of the multilayer perceptron is again a vector, containing the classification values for the shapes \emph{rectangular} and \emph{circular}. It is converted to percentages as well.

Both classifiers have been thoroughly tested. The following numbers give the evaluation results for the color test. 
Hereby, $Y(o)$ yields the classification result with an object \emph{o} as input and $X(o)$ returns the real feature of the object. 

\begin{table}[h!]  
		\centering 				
		\begin{tabular}[h!]{ccc} 		
		$P(Y(o)~=~red~ \vert ~X(o)~=~red)$ & $\approx$ & 0.82\\
		$P(Y(o)~=~green~ \vert ~X(o)~=~green)$ & $\approx$ &0.86 \\
		$P(Y(o)~=~yellow~ \vert ~X(o)~=~yellow)$ & $\approx$ & 0.96\\
		$P(Y(o)~=~brown~ \vert ~X(o)~=~brown)$ & $=$ & 1\\\\
		$P(X(o)~=~red~ \vert ~Y(o)~=~red)$ & $=$ & 1\\
		$P(X(o)~=~green~ \vert ~Y(o)~=~green)$ & $\approx$ & 0.93 \\
		$P(X(o)~=~yellow~ \vert ~Y(o)~=~yellow)$ & $=$ &1\\
		$P(X(o)~=~brown~ \vert ~Y(o)~=~brown)$ & $\approx$ & 0.77
		\end{tabular}
\end{table}

Thus, $P( Y(o)=a ~ \vert ~ X(o)=a )$ gives the conditional probability that the system classifies the color of an object 
\emph{o} to be \emph{a}, under the condition that the color was \emph{a}. 
$P( X(o)=a ~ \vert ~ Y(o)=a )$ on the other hand gives the conditional probability that an object has 
the color \emph{a}, given that the system classified it as having color \emph{a}.

As the probabilities show, most misclassifications happened with red objects, which were mostly mistaken for brown ones. The same holds for green objects. This is reflected in the last probability, showing that if the system reported an object to be brown, this was only correct in 77 percent. All together this still gives a color classification that is sufficiently reliable for our purposes.

In the following the conditional probabilities for the shape evaluation are given -- the abbreviation \emph{rect.} stands for \emph{rectangular}. As can be noticed, that the reliability is even higher than that of the color classification. Hence, this classification is sufficient for our purpose, too. Both classifiers are learned offline, i.e. before the robot is presented objects.

\begin{table}[h!]  
		\centering 				
		\begin{tabular}[h!]{ccc} 		
	$P(Y(o)~=~circular~ \vert ~X(o)~=~circular)$ &=& 1\\
	$P(Y(o)~=~rect.~ \vert ~X(o)~=~rect.)$ & = & 0.97\\\\
	$P(X(o)~=~circular~ \vert ~Y(o)~=~circular)$ & $\approx$ & 0.97 \\
	$P(X(o)~=~rect.~ \vert ~Y(o)~=~rect.)$ & = &1
		\end{tabular}
\end{table}

\subsection{Knowledge Management}

The knowledge management unit is one of two major components of the system. All learned knowledge is administered here, including the analogy making and knowledge generalization. Thus, the actual action selection in response to a presented object is performed in this component, too (cf. figure \ref{fig:wholeSystem}). The first subsection briefly describes how the knowledge is represented. This is followed by an overview of the algorithm that selects actions and evolves the knowledge base.

\subsection*{Knowledge Representation}

As has been mentioned before, the knowledge is represented symbolically by a graph. This graph contains the data of all known object categories, including the features linked to each object category and the experiences the agent has made with it as well as its similarity to other object categories. 

\begin{figure}
\begin{center}
\resizebox{8.8cm}{3.8cm}{
\begin{tikzpicture}[
		>=stealth',
		auto,
		node distance=3cm,
		thick,
		cell/.style={rectangle,draw=black},
		id node/.style={circle,fill=gray!50, draw, double,font=\sffamily\Large\bfseries},
		attribute node/.style={rectangle, draw, fill=gray!20, text width=5em, text centered, rounded corners},
		experience node/.style={rectangle, draw, text width=5em, text centered, rounded corners}
	]

	\coordinate (origin) at (0,1) ;
	\node[attribute node] (Color) at (-5.8,-1.5) {Color};
	\node[attribute node] (Form) at (0,-1.5) {Form};
	\node[attribute node] (Experience) at (5.8, -1.5) {Experience};
	\node[experience node] (Action1) at (4.5,-5) {Action1};
	\node[experience node] (positive)  at (7,-5)  {positive};
	\node[id node] (id) at (origin) {$s_3$};

	\node [draw, rounded corners] (color1) at (-7.5, -5) {
		\begin{tabular}{c|c|c}
			red & 0 & 0 \\
			\hline
			green & 1 & 1 \\
			\hline
			yellow & 0 & 0 \\
			\hline
			brown & 0 & 0 \\
		\end{tabular}
	};

	\node [draw, rounded corners] (color2) at (-4, -5) {
		\begin{tabular}{c|c|c}
			red & 0.7 & 0.7 \\
			\hline
			green & 0 & 0 \\
			\hline
			yellow & 0 & 0 \\
			\hline
			brown & 0.3 & 0.3 \\
		\end{tabular}
	};

	\node [draw, rounded corners] (form1) at (0, -5) {
		\begin{tabular}{c|c|c}
			circular & 0 & 0.2 \\
			\hline
			rectangular & 0.8 & 1 \\
		\end{tabular}
	};

	\path[every node/.style={font=\sffamily\small}]
		(id) edge (Color)
		(id) edge (Form)
		(id) edge (Experience)
		(Experience) edge (Action1)
		(Experience) edge (positive)
		(Color) edge node [right] {\LARGE $\frac{1}{3}$} (color1)
		(Color) edge node [right] {\LARGE$\frac{2}{3}$} (color2)
		(Form) edge node [right] {\LARGE$\frac{3}{3}$} (form1);

\end{tikzpicture}
}
\caption{An example knowledge graph containing data of one object category}
\label{fig:exampleGraph}
\end{center}
\end{figure}
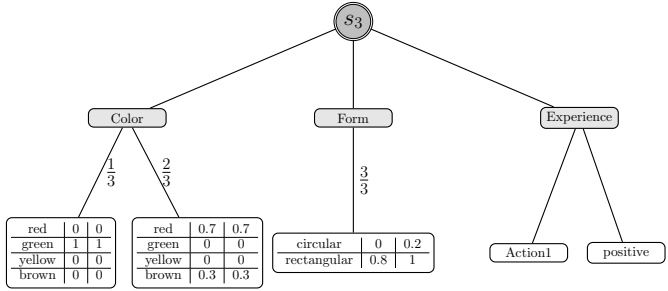

\mybox{An example graph containing only one object category can be seen in figure \ref{fig:exampleGraph}. The object category in this example is labeled as $s_3$ and has the color features 

\[[re: 0.7-0.7; gr: 0-0; ye: 0-0; br: 0.3-0.3]\]\\
as well as 

\[[re: 0-0; gr: 1-1; ye: 0-0; br: 0-0]\] \\
attached to it. Hereby \emph{re} denotes \emph{red}, \emph{gr} \emph{green}, \emph{ye} \emph{yellow} and \emph{br} denotes \emph{brown}. The entry for each color characteristic -- e.g. \emph{red} --  is an interval.}

The feature intervals connected to an object category are needed to determine whether a perceived object fits into an object category. An object is defined as fitting into a category if for each of its feature vectors an appropriate feature vector can be found whose characteristic intervals all contain the characteristic values of the object. Please note that this must hold for all features. 

\mybox{In the example graph an object with the color 

\[[0.7, 0, 0 ,0.3]\]\\
and the form 

\[[0.1, 0.9]\]\\
 would fit into the category $s_3$. But an object with the color 

\[[0.2, 0, 0 ,0.8]\]\\
 and the same form would not fit into the category, since for the color vector no vector associated with the category can be found, such that all characteristics lie within the characteristic intervals of the vector.} 

The links connecting the feature vectors to the name of the corresponding feature have an occurrence probability.

\mybox{In the example case this means that so far three objects of the category $s_3$ have been encountered, of which two were to 70 percent red, the third green in color. Furthermore, objects of this category have a form that has been classified as  0-20 percent circular vs. 80-100 percent rectangular.

As experience a \emph{positive} reward for \emph{Action1} has been obtained. In the example scenario this could be an object category for red and green toy blocks where \emph{Action1} denotes the sorting into the toy box.}

\subsection*{Knowledge Management Algorithm}

Whenever the system detects an object, the extracted feature vectors are relayed to the knowledge management component. Here it is checked whether the reported object already fits into an object category as described above.

In case the detected object fits into a category, the respective feature vector probabilities as well as the similarity of this object category to all others are updated. Afterwards, an appropriate action is selected. How such a similarity is calculated will be explained shortly. 

If the detected object does not fit into any known object category, a new object category is created. Then the similarity of the new category to all other categories is calculated and an appropriate action is chosen. Next the action selection process is described.

\subsubsection*{Action selection}

An action is selected based on previous experiences according to the following scheme.

\begin{itemize}
\item Choose an action that has a positive experience with the present object category. 
\item If such an action does not exist, the object categories with a positive similarity to the current category are checked for positive experiences, starting with the most similar category. Nonetheless, the action selected should not have received a negative reward with either the current or the most similar object category. 
\item If such an action still does not exist, a random action is chosen from the set of actions that have no negative experience with the current object category. 
\end{itemize}

Instead of using the above presented scheme to determine an action, a random action is sometimes chosen to avoid staying in local maxima -- the probability is given by the parameter $\rho_{ra}$. 

\begin{table}[h]  
		\centering 				
		\begin{tabular}[h!]{cc} 		
		\multirow{2}{*}{$\rho_{ra}$ :} & Probability of choosing\\
		& a random action
		\end{tabular}
\end{table}

\mybox{Suppose that in the example scenario the robot has only encountered red apples and toy blocks of various colors so far. Now it is presented a green apple. This green apple does not fit into any of the known ca-tegories, consequently no experiences exist with it. But by comparing its features to the known categories \emph{red apples} and \emph{toy blocks}, it could detect a high similarity with the \emph{red apples} category. The robot has made positive experiences with sorting red apples into the fruit basket. Therefore, this action is chosen for the green apple, too.}

\subsubsection*{Similarities Between Object Categories}

In the above description, similarities between object categories have been mentioned multiple times. According to \cite{Tversky} not only similar, but also differing features should be taken into account for such a similarity measurement. \cite{Luger} further mentions that not all features should be weighted equally. As a consequence, it is proposed that the similarity between two object categories is calculated by summing up the weighted similarities regarding single features as well as the weighted similarity regarding experience. 

Therefore, let the following symbols be given:
			
	\begin{tabular}[h!]{rl} 		
	$M$ & Number of features\\
	$s_j, s_k$ & Two object categories\\
	$\omega_i$ & Weight of feature $i$\\
	$\omega_e$ & Weight for the experience\\
	$\sigma_{f_i, s_j, s_k }$ & Similarity of the object categories $s_j$\\ 
	 & and $s_k$ regarding feature $f_i$ \\
	$\sigma_{e, s_j, s_k }$ & Similarity of the object categories $s_j$\\
	&  and $s_k$ regarding the experience\\
	\end{tabular}

Then the \emph{similarity between two object categories} $s_j$ and $s_k$ is computed with equation \eqref{e1}.

\begin{equation} \label{e1}
\sigma_{s_j, s_k} = \sum_{i=1}^{M}{\omega_i \cdot \sigma_{f_i, s_j, s_k}} + \omega_e \cdot \sigma_{e, s_j, s_k } 
\end{equation}

The similarity of two object categories regarding a single feature (\emph{feature similarity}) is calculated using equation \eqref{e2}. 

\begin{equation} \label{e2}
\sigma_{f, s_j, s_k}~ = ~\sum_{c_j\in C_j} \max\limits_{c_k \in C_k}\bigg( (1-\Delta_{jk}) \cdot P(c_j) \cdot P(c_k) \bigg)
\end{equation}

In this equation, $C_j$ and $C_k$ are the sets of associated feature vector intervals of feature $f$ for the object categories $s_j$ and $s_k$. The cardinality of $C_j$ is always smaller or equal to the cardinality of $C_k$. If this is not the case, $s_j$ and $s_k$ are interchanged. This guarantees that the similarity of two object categories is not dependent on the order in which they are compared to each other -- i.e. $s_j$ compared to $s_k$ produces the same result as $s_k$ compared to $s_j$. $P(c_i)$ represents the occurrence probability of the feature vector interval $c_i$ for the object category $s_i$. $\Delta_{jk}$ on the other hand encodes the overall distance of the compared feature vector intervals by summing up the shortest difference between the intervals for each characteristic of the feature. 

\mybox{As an example the feature similarity between the following feature interval vectors is calculated.

\[[re: 0.7-0.7;~ gr: 0-0;~ ye: 0.3-0.3;~ br: 0.1-0.1]\]
\[[re: 0.6-0.8;~ gr: 0-0;~ ye: 0-0;~ br: 0.3-0.4]\]\\
The two \emph{red} (re) characteristic intervals have a distance of 0, since they overlap. The same holds for the \emph{green} (gr) characteristic intervals. The intervals for the characteristic \emph{yellow} (ye) on the other hand do not overlap, but have a distance of 0.3; the distance for the \emph{brown} (br) characteristic is $0.3 - 0.1 = 0.2$. By summing up these values up, a value of 0.5 is received for $\Delta$.} 

The value of $\Delta$ has a range of [0;2], whereby smaller values represent a greater similarity. To make this value more intuitive, it is mapped to [-1;1] in which negative values represent a dissimilarity and positive values a similarity. This is achieved by subtracting $\Delta$ from 1 (see equation \eqref{e2}). In order to ensure that the similarity measurement values common feature vector intervals more highly than uncommon ones, the occurrence probabilities of the feature vector intervals are taken into account as well (see equation \eqref{e2}). These occurence probabilities are encoded in the knowledge graph, as can be seen in figure \ref{fig:exampleGraph}.

\subsubsection*{Dynamic Weight Adaption}

As mentioned before, different features as well as the experience are weighted differently for the similarity calculation. The \emph{weights} -- named $\omega_i$ and $\omega_e$ in equation \eqref{e1} -- all have an initial value of 1 and are dynamically adapted at run time. At all times all weights are greater or equal to zero and the sum of all weights is constant. An adaption of weights occurs in three different cases:

\begin{itemize}
\item Two object categories are merged.
\item An object category is split into two categories.
\item An experience is made for the first time for a given category-action pair and it contradicts the experience of the most similar object category.
\end{itemize}

In the first case, the two object categories are thought of as very similar. Therefore, the weights should make the detection of similarities of the same type easier. This is achieved by decreasing the weight and therefore the importance of that attribute (feature or experience) in which the two categories differ the most. All other weights are increased. In the second case, the resulting two object categories are apparently not similar although the similarity calculation implies that. Consequently, the weight of the attribute that is the most responsible for their alleged similarity is decreased and all other weights are increased. The same holds for the third case. By how much the weights are decreased or increased at a time is controlled by the parameter $\delta_{aw}$.

\begin{table}[h]  
		\centering 				
		\begin{tabular}[h!]{cc}
		\multirow{2}{*}{$\delta_{aw}$ :} & Controls by how much weights\\
		&  are in- or decreased
		\end{tabular}
\end{table}

\subsubsection*{Reward Processing and Knowledge Generalization}

So far the overall algorithm has chosen an action as a response to a presented object. This action is executed in the next step and a reward is given by a human operator. This experience is then automatically added to the knowledge base. If such an experience already exists in the graph and both -- the saved and  the newly received reward -- are the same, no update is needed. If one of the rewards is \emph{neutral}, the other reward is saved. However, if one reward is \emph{positive} and the other \emph{negative}, this is perceived as a strong indicator that the object does not fit the object category. The currently presented object is then \emph{split} from the object category to form a new category. In any case the similarity values between object categories have to be updated again.

Afterwards as long as object categories exist whose similarity is greater than or equal to a threshold $\theta_{mc}$, these object categories are \emph{merged}. During a merging operation feature vector intervals can also be merged if they have a distance $\Delta$ (cf. above) smaller than or equal to another threshold $\theta_{mf}$. After a merging operation, similarity values between object categories have to be recalculated again. 

\begin{table}[h]  
		\centering 				
		\begin{tabular}[h!]{cc} 		
		\multirow{2}{*}{$\theta_{mc}$ :} & Threshold, controls \\
		 & merging of categories\\\\
		\multirow{2}{*}{$\theta_{mf}$ :} & Threshold, controls\\	
		& merging of feature vector intervals
		\end{tabular}
\end{table}

\mybox{In the example, the agent has chosen to sort the green apple into the fruit basket. After the execution, it receives a positive reward for this action from a human supervisor. This experience is then entered into the knowledge base. Since no previous experience existed, splitting is not considered. Furthermore, the experience of the most similar object category -- \emph{red apples} -- with this action is not contradictory, therefore no weight adaption is needed either. Yet the similarities to all other object categories have to be updated to take the new experience into account. As a result, the similarity with the object category \emph{red apples} is larger than the threshold $\theta_{mc}$. Hence, the two object categories are merged. The two merged categories differed the most in the color feature. As a result, the weight of the color feature is decreased by the value of $\delta_{aw}$ for the similarity calculation and the weights for color and experience are each increased by the value of $\delta_{aw}/2$. Consequently, the similarities between all object categories have to be updated again. If no other object categories need to be merged, the robot is ready to interact with another object.}

Overall, the results of the algorithm are controlled by four parameters. The parameter $\rho_{ra}$ controls how often random actions are chosen. $\delta_{wa}$ defines the amount by which the weights for the different attributes are decreased or increased for the simularity calculation. The threshold parameters $\theta_{mc}$ and $\theta_{mf}$ regulate the merging of categories and feature vectors respectively. How the parameter settings affect the results is discussed in the next section.

\section{Evaluation of the proposed framework}

The presented system was evaluated with two different scenarios. The first was the example scenario mentioned in section \ref{sec:example}. This scenario was also used to analyze the effects the parameters have on the system's performance.
The second tested scenario was the so called \emph{Wisconsin Card Sorting Test} (WCST), which will be explained later.

\subsection{Example Scenario}

To evaluate the system's performance for the example scenario, it was analyzed after how many presented objects the desired object categories had been formed and were not changed anymore. For the example scenario, the desired outcome consists of three object categories: one for both green and red apples, one for toy blocks and one for brown apples. The system was tested using real world objects. However, to prevent feature extraction problems to alter the test, the scenario was additionally tested with simulated feature vectors in two variants: In one variant, the simulated features were 100 percent correct, in another the features were classified correctly but contained uncertainties.

\mybox{An example of a simulated, 100 percent correct color feature vector is given as follows:

\[[re: 1;~ gr: 0;~ ye: 0;~ br: 0]\]\\
An example for a simulated color feature vector containing uncertainties on the other hand is given as:

\[[re: 0.8;~ gr: 0;~ ye: 0.05;~ br: 0.15]\]}

In all three variants of the scenario parameter settings could be found for which the system reached the desired outcome very quickly. A lower bound for the number of steps it takes the system to acquire all requested object categories is of course the number of objects shown to the system before it has seen an object of every category. With some parameter settings the system was able to reach this lower boundary, meaning that it had formed the desired object categories as soon as it had seen at least one object of every category.

The order in which objects were presented to the system was also permuted, yet it influenced the system's performance only slightly. During the real world tests a few misclassifications occurred. However, the system was only temporarily 'confused' and managed to reach the desired outcome none the less. As has been mentioned briefly, the chosen parameters have an influence on the system's performance. In order to evaluate the impact of two of the four parameters, the example scenario was tested in all three variants with different parameter settings.

\begin{figure}
\begin{center}
\includegraphics[angle=0,width=8.4cm]{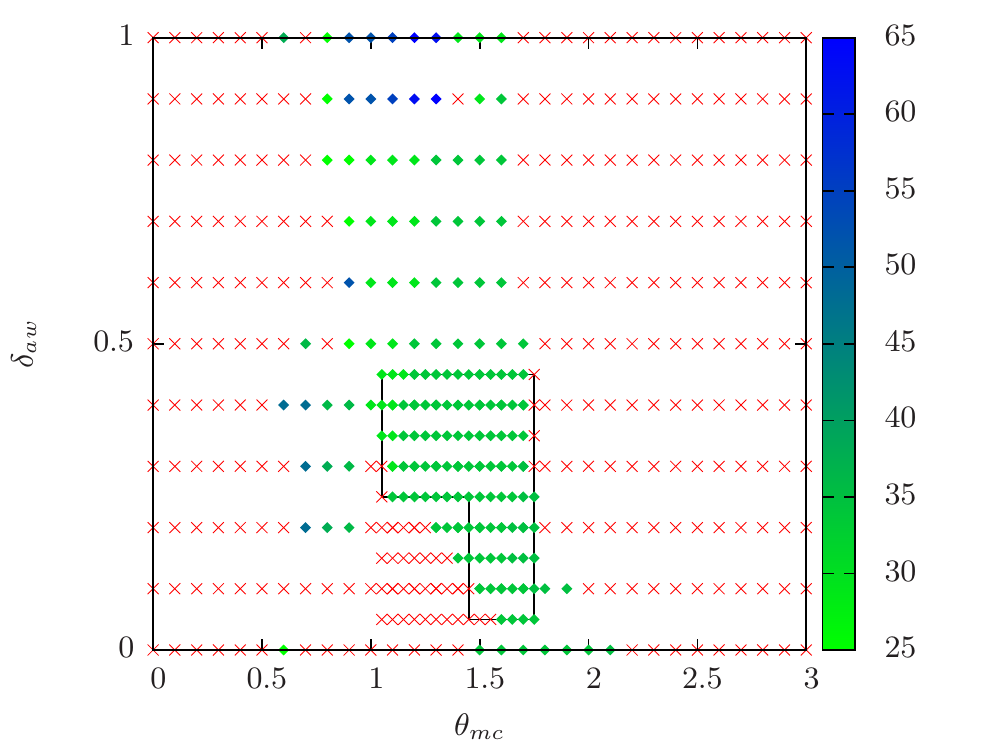}    
\caption{System performance for different parameter combinations}  
\label{fig:parametersensitivity}                                 
\end{center}                                 
\end{figure}

Figure \ref{fig:parametersensitivity} shows the results for variations of the threshold parameter ($\theta_{mc}$) and the weight adaption parameter ($\delta_{aw}$). The other two parameters remained constant during the test with values of $\rho_{ra}=0$ and $\theta_{mf}=0.3$. Red crosses indicate that the desired object categories were not reached within the allotted time frame; dots indicate that the desired object categories were formed, the color encoding the number of objects that had to be presented before the state was reached. The encircled area contains parameter combinations which produced good results in all test.
It has to be noted that only after approximately 25 objects had been presented to the robot, objects of all three categories had been shown. Therefore, the desired object categories could only be found after 25 steps. As can be seen in figure \ref{fig:parametersensitivity}, the best parameter combinations reach this state slightly after 25 steps and are therefore very fast.

Overall the parameter $\theta_{mc}$ seems to have a larger influence on the system performance than the $\delta_{aw}$ parameter. However, the parameter combinations that produce good results form a large area, meaning that slight parameter variations can be tolerated.

\subsection{Wisconsin Card Sorting Test}

\begin{figure}
\begin{center}
\includegraphics[angle=0,width=6cm]{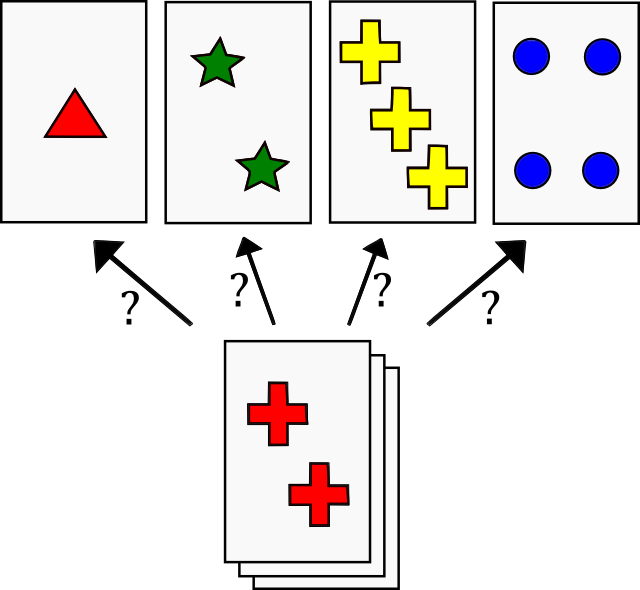}    
\caption{The setting of the \emph{Wisconsin Card Sorting Test}}  
\label{fig:wcst}                                 
\end{center}                                 
\end{figure}

The system's relearning abilities were evaluated using the \emph{Wisconsin Card Sorting Test} by \cite{Berg}. In this test the subject is shown four cards. One card shows one red triangle, another two green stars. A third card depicts three yellow crosses and a fourth four blue circles. The subject is then given a set of 60 cards, each displaying up to four objects. One a single card all objects have the same shape and the same color from the set of four colors and four shapes mentioned above. The subject is then asked to assign each card to one of the four other cards. This setting is depicted in figure \ref{fig:wcst}.

After every assignment the proband is told if the action was correct. Unknown to the subject, he has to sort the cards either according to color, form or number of shown objects. However, whenever he assigns five cards correctly in a row, the rule which determines whether an assignment is correct is changed. The test ends when the subject assigned nine times five cards in a row correctly. If this has not been accomplished after 60 cards, the cards are mixed and given back to the proband. 

This test was performed with human subjects by \cite{Berg}. Fifty percent of the subjects managed this test quite well. However, the rest of the test group was not able to complete the test. These subjects especially had problems with relearning. Despite receiving negative rewards after the assignment rule changed, they kept choosing previously correct actions. For these subjects, the test had to be aborted after some time.

The relearning abilities of the proposed system were evaluated by performing this test using simulated feature vectors. Unlike the human subjects, the robot was able to complete the test at all times, yet the time required to relearn was far longer than for the human subjects that completed the test. Additionally, these human subjects were sometimes able to predict the rule changes, thereby allowing them to react more quickly to them. But since the system was not designed to keep track of such changes, the robot was unable to do so. Though when looking at the similarity weights for the different features, it can be seen that at a time directly preceding a rule change, the weight of the feature to use as a sorting rule is quite higher than the other weights. This implies that the system correctly identified the underlying sorting rules.

\section{Conclusion}

Modern artificial intelligence systems should use a hybrid knowledge representation. Furthermore, they should take the symbol grounding problem into account. The presented system fulfills both requirements. It offers a flexible way to learn object 
categories and behavior patterns in unpredictable environments and allows the robot to form analogies to react to new objects appropriately. The symbol grounding problem is avoided by embedding the sensory information directly into the symbolic knowledge representation.
 
The system has been tested both in real and in simulated environments. The algorithm's parameters have been shown to influence the results, yet a large group of parameter settings exists for which good results could be achieved in all tested scenarios.

Yet the system also makes assumptions that might be too restrictive. For once it assumes that the similarity between objects is calculated the same way no matter what objects are involved. As a result it is assumed that all objects can be grouped in a similar way. Nonetheless, in real world scenarios this might not be the case. The experience factor for the calculation of 
the similarity between object categories can balance this, but only at the cost of feature importance.
Furthermore, the way object categories are formed heavily depends on the extracted features. If the key for object classification lies in a feature that is not detected, no reasonable object categories can be formed. As such, the applicability
of the system needs a feature detection that is as diverse as possible.

While performing the tests, different misclassifications occurred. These errors often led to the formation of isolated
object categories that were never merged with other categories. As a result, the graph grew unnecessarily large and the system could no longer reach the desired state of having exactly the predicted object categories. Nonetheless the actions selected by the system were unaffected by these additional object categories. To prevent the long time persistence of erroneous object categories in the future, a decay of knowledge that relies on a single occurrence could be introduced.

Another flaw of the presented system is the reward model which only anticipates 
\emph{positive}, \emph{negative} and \emph{neutral} rewards. Additionally it assumes that the reward for an object
is always the same. To overcome this flaw, an advanced model of rewards such as that of so called \emph{somatic markers} by \cite{Hoefinghoff}, \cite{Hoefinghoff13} could be used which would also allow varying rewards.

Besides solving the above mentioned problems and too restrictive assumptions, future work might concentrate on the incorporation of multiple knowledge graphs, one for each task to be performed by the robot. This is necessary since the reward for a single action highly depends on the task that shall be performed. The same holds
for the formation of object categories. Imagine different scenarios in which a robot shall choose objects for
decoration purposes. The action \emph{choose red apple} might receive a positive reward when decorating a 
red environment, while the same action for the same object might be punished if a blue environment shall be decorated.
To solve this problem, a separate knowledge graph could be formed for every task. But since knowledge from one task 
might be reused in another task, the different graphs should be interconnected. This means that whenever 
a new object occurs for a certain task, experiences with this object from a different task could be used.

\end{document}